\documentclass[11pt]{article}
\usepackage[preprint]{acl}
\usepackage{times}
\usepackage{latexsym}
\usepackage{amsmath,amssymb}
\usepackage{booktabs}
\usepackage{microtype}
\usepackage[table]{xcolor}
\usepackage{graphicx}
\usepackage{multirow}
\usepackage{url}
\usepackage{float}
\usepackage{placeins}
\usepackage{stfloats}
\usepackage{algorithm}
\usepackage{algpseudocode}

\newtheorem{proposition}{Proposition}

\newcommand{\method}{\textsc{HSPR}}

\newcommand{\E}{\mathbb{E}}

\title{HSPR: Recovering Software-Agent Benchmark Scores\\
via Hierarchical Posterior Inference}

\author{%
  Wanyi Chen\textsuperscript{1}\quad
  Wanqin Shen\textsuperscript{1}\quad
  Yuanchen Shi\textsuperscript{1}\quad
  Zhensu Sun\textsuperscript{2}\quad
  Fang Kong\textsuperscript{1}\thanks{Corresponding author.} \\
  \textsuperscript{1}Soochow University \\
  \textsuperscript{2}Singapore Management University
}

\hypersetup{
  pdftitle={HSPR: Recovering Software-Agent Benchmark Scores via Hierarchical Posterior Inference},
  pdfauthor={Wanyi Chen, Wanqin Shen, Yuanchen Shi, Zhensu Sun, Fang Kong}
}

\begin{document}
\raggedbottom
\maketitle

\begin{abstract}
Evaluating software agents on a full benchmark is expensive. A small
task subset can preserve agent rankings while yielding inaccurate scores.
Historical responses provide useful structure, but tasks with identical
source responses can have different outcomes for a new target agent.
We introduce \textbf{H}ierarchical \textbf{S}core \textbf{P}osterior
\textbf{R}ecovery (\textbf{HSPR}), which combines
source-fitted response factors with uncertain group success rates.
Observed target outcomes update the posteriors over the target latent
state and group rates; posterior-guided acquisition selects subsequent tasks. Its score-based
criterion equals the expected one-step reduction in full-score posterior
variance. Observed successes and posterior-predictive counts for unqueried
tasks yield a full-score posterior. The final estimate uses its median
with source-selected calibration. At a 10\% task budget on four task banks
from the SWE-bench family, HSPR achieves the lowest mean absolute error
(MAE) on three banks and the lowest maximum absolute error (MaxErr) on
all four among the methods compared. Removing group uncertainty or
replacing adaptive queries increases macro MAE in the respective controls.
Macro MAE decreases across 5--20\% budgets.
Code and project resources are publicly available.\footnote{%
Project page: \href{https://wanyichen06.github.io/HSPR/}{\nolinkurl{wanyichen06.github.io/HSPR/}}.\\
Code: \href{https://github.com/wanyichen06/HSPR}{\nolinkurl{github.com/wanyichen06/HSPR}}.}
\end{abstract}

\begin{figure}[t]
\centering
\includegraphics[width=\columnwidth]{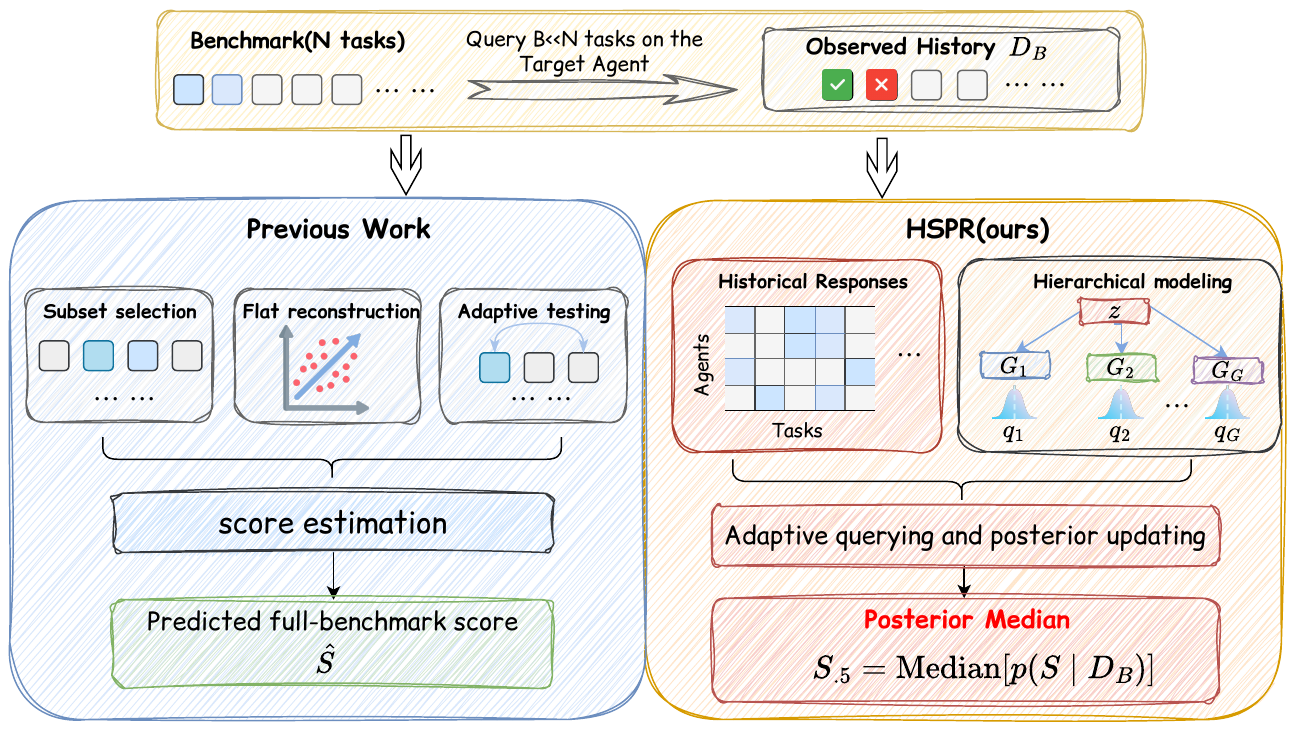}
\caption{Task-efficient software-agent evaluation. Representative strategies
estimate full-benchmark scores from limited target executions. HSPR combines
hierarchical response modeling with adaptive querying. The displayed median
is the decision before source-selected calibration.}
\label{fig:intro-overview}
\end{figure}

\section{Introduction}
Evaluating a software agent requires repository exploration, code editing,
tool use, and execution across hundreds or thousands of tasks
\citep{jimenez2024swebench,duan2026pta}. A small execution budget makes
this evaluation more practical and creates an inference problem:
recovering the agent's score on the \emph{full benchmark} from a few
observed task outcomes. Accurate score recovery is essential for
comparing progress and assessing capability.

At a 10\% budget on SWE-bench Pro, Mid-Range selection
\citep{ndzomga2026efficient} preserves model ordering well
($\tau_b=.967$), yet its raw subset mean has mean absolute error (MAE)
of 22.798 percentage points (pp) (Table~\ref{tab:low-cost-leaderboard}).
Using the \emph{same Mid-Range outcomes}, a source-fitted Ridge
estimator reduces MAE to 3.434 pp. Estimation accuracy also varies
across task banks: ATLAS-port has MAE of 0.957 pp on Full and
7.736 pp on Pro. Low-budget evaluation depends on both task selection
and the estimator applied to the observed outcomes.

The source-response signature of a task records its binary outcomes
across historical agents. Tasks with identical signatures may have
different outcomes for a new target agent, while still sharing a response
tendency. Modeling this tendency as an uncertain group success rate lets
one observation inform other group members without fixing their outcomes.
We ask:
\emph{How can we recover a target agent's full-benchmark score from a few
executions while accounting for uncertainty in how source response patterns
transfer to the target?}

We introduce \textbf{H}ierarchical \textbf{S}core \textbf{P}osterior
\textbf{R}ecovery (\textbf{HSPR}), illustrated in Figure~\ref{fig:intro-overview}.
HSPR uses source-fitted response factors to model a target latent response
state and uncertain success rates within source-signature groups.
Each observed outcome updates the latent-state posterior and the group-rate
posteriors. Posterior-guided acquisition selects the next task using these
dependencies. Observed successes and posterior-predictive counts for
unqueried tasks yield a distribution over the full-benchmark score.
Its median minimizes model-conditional expected absolute loss; source-only
validation selects calibration for the final estimate.

At a 10\% task budget, HSPR achieves the lowest MAE point estimates on
Lite, Full, and Pro and the lowest MaxErr on all four SWE-bench task banks
compared with the baselines in Table~\ref{tab:low-cost-leaderboard}.
On Verified, its MAE is within 0.055 pp of ATLAS-port.
In the respective controls, removing group uncertainty increases macro
MAE, as does replacing adaptive queries with the estimator held fixed.
Across 5--20\% task budgets, equal-bank macro MAE decreases as more
outcomes are observed.

Our contributions are threefold:
\begin{itemize}
\item We introduce HSPR, a hierarchical full-score model that combines
source-fitted response factors with uncertain target success rates
within source-signature groups.
\item We derive posterior-guided acquisition for score recovery,
including a criterion equal to the expected one-step reduction in
finite-score variance. Source-only validation selects the deployed
acquisition objective before target evaluation.
\item We provide a controlled empirical study across four SWE-bench
task banks, including end-to-end hierarchy controls, estimation and
acquisition ablations, paired statistical analysis, mechanism diagnostics,
and 5--20\% budget sweeps.
\end{itemize}

\begin{figure*}[t]
\centering
\includegraphics[width=\textwidth,trim=0 90bp 0 0,clip]{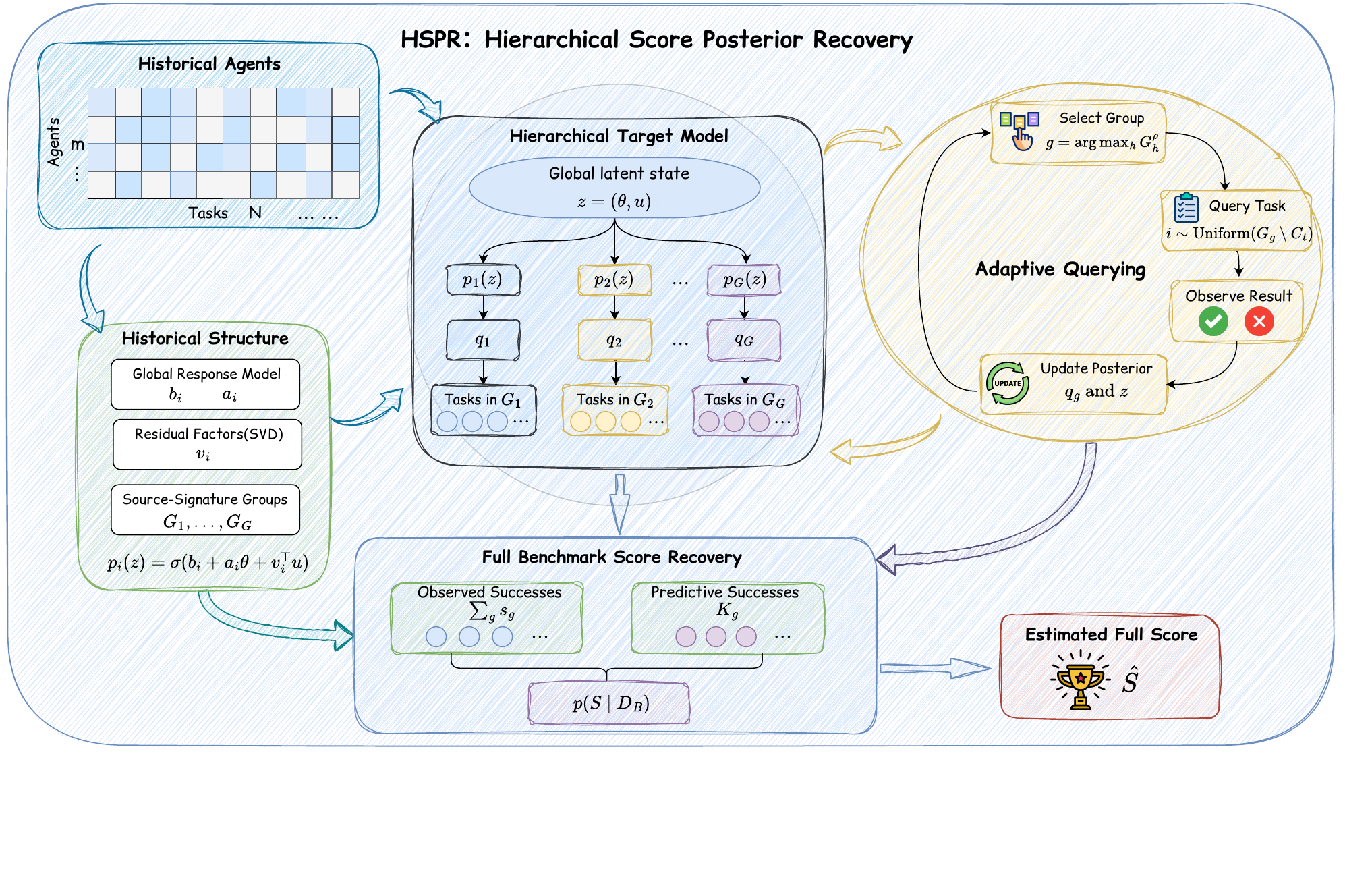}
\caption{HSPR evaluation pipeline. Source responses provide fitted
response factors and source-signature groups. Sequential queries update
the posteriors over the target latent state and group success rates.
Observed successes and posterior-predictive success counts for unqueried
tasks define the full-score posterior used for the final calibrated estimate.}
\label{fig:hspr-method-overview}
\end{figure*}

\section{Related Work}
\label{sec:related-work}
\paragraph{Efficient benchmark prediction.}
Active Testing combines informative queries with sampling correction
\citep{kossen2021active}; Active Surrogate Estimators learn a surrogate
for unobserved errors \citep{kossen2022surrogate}.
Random-Sampling-Learn (RSL) uses random-core regression; learned benchmark
predictors can fail under extrapolation \citep{zhang2025benchmark}.
tinyBenchmarks uses compact calibrated subsets \citep{polo2024tiny},
while SparseEval optimizes anchors and weights from the response matrix
\citep{zhang2026sparseeval}.
Factorized Active Querying (FAQ) combines historical factors with adaptive
querying and finite-population inference \citep{wu2026faq}.
HSPR uses a model-conditional finite-score posterior with global
factors and uncertain group rates.

\paragraph{Psychometric evaluation.}
Item-response models calibrate NLP test items \citep{lalor2016scale}.
Amortized model-based evaluation predicts item difficulty from question
content to support adaptive testing \citep{truong2025amortized}.
Agent Psychometrics models task-level agent responses
\citep{ge2026psychometrics}. ATLAS uses item-response models with
Fisher-guided acquisition \citep{li2026atlas}; PTA-IRT uses historical
trajectory information for task selection and reconstruction
\citep{duan2026pta}. HSPR builds on historical item calibration but adds
residual response dimensions and conditional target-specific group rates.
ATLAS's Fisher-guided acquisition targets latent ability.
HSPR's score-variance criterion targets the full-benchmark score,
accounting for dependence among unobserved outcomes.

\paragraph{Subsets and objectives.}
Historical pass-rate selection can preserve software-agent rankings
\citep{ndzomga2026efficient}; subset quality depends on task banks and
representations \citep{gusev2026ranking}.
SubLIME predicts subset-to-full rank correlation from anchor models and
dataset characteristics \citep{saranathan2025sublime}.
Ranking measures relative ordering
\citep{liu2025amelo}, while score recovery requires an accurate numerical
scale.

\paragraph{Agent evaluation context.}
HELM, BIG-bench, MMLU, and Dynabench cover heterogeneous evaluation
settings \citep{liang2022helm,srivastava2022beyond,hendrycks2021mmlu,kiela2021dynabench}.
AgentBench and AgentBoard evaluate tool use and multi-turn agents
\citep{liu2023agentbench,ma2024agentboard}.
Agent$^2$ RL-Bench evaluates agents on end-to-end reinforcement-learning
post-training engineering \citep{chen2026agent2rlbench}.
EarlyEval reduces within-task execution through early outcome prediction
\citep{shi2026earlyeval}.
Trajectory-based selection \citep{ayyad2026sail} and process-aware adaptive
testing \citep{vanderlinden2008response} use additional observations.
HSPR uses source binary responses and observed target outcomes.

\section{Hierarchical Score Posterior Recovery}
\label{sec:method}
Figure~\ref{fig:hspr-method-overview} summarizes HSPR from historical response
modeling to adaptive target queries and full-score recovery.
Algorithm~\ref{alg:hspr} specifies the evaluation procedure.
\subsection{Finite-Score Estimation}
Let $M\in\{0,1\}^{m\times N}$ be the source response matrix of $m$ agents
on a fixed bank of $N$ tasks. The target outcome vector
$Y\in\{0,1\}^N$ defines the full-benchmark score
\begin{equation}
 S=\frac{1}{N}\sum_{i=1}^N Y_i.
 \label{eq:score}
\end{equation}
The evaluator observes at most $B$ distinct target-task outcomes and
returns an estimate $\widehat S$ of $S$. Let
$C_t=\{i_1,\ldots,i_t\}$ be the queried task set and
$D_t=\{(i_j,Y_{i_j})\}_{j=1}^t$ the observed history.
In archived replay, $Y$ is fixed and only queried entries are revealed.
HSPR models the unobserved entries probabilistically. We condition on
$M$ throughout; acquisition at step $t$ uses only $M$ and $D_t$.

The score separates into observed and unobserved outcomes:
\begin{equation}
 NS=\sum_{i\in C_t}Y_i+\sum_{i\notin C_t}Y_i.
 \label{eq:partition}
\end{equation}
The first term is exact. HSPR places a joint predictive distribution on
the second, preserving dependence among unobserved outcomes.

\subsection{Historical Factors and Groups}
\paragraph{Source response representation.}
We fit a regularized logistic item model on historical agents. For task $i$,
$b_i$ is an intercept and $a_i$ an ability slope. We obtain source-agent
ability values by standardizing the logits of their smoothed full-benchmark
scores. From source responses, we fit a prior over task success rates
that mixes a point mass at zero with a Beta distribution. It supplies initial probabilities
used to initialize and regularize the item intercepts.

After item fitting, a centered SVD of the source response residuals
provides $r$ task directions. Scaled residual loadings $v_i\in\mathbb R^r$
describe departures from the scalar ability pattern. The source-fitted
item parameters $(b_i,a_i,v_i)$ are fixed during target inference.
The target latent response state $z=(\theta,u)\in\mathbb R^{1+r}$ combines
scalar ability $\theta$ with residual coordinates $u$:
\begin{equation}
 z\sim\mathcal N(0,P^{-1}),\qquad
 p_i(z)=\sigma(b_i+a_i\theta+v_i^\top u).
 \label{eq:global}
\end{equation}
Here $\sigma$ is the logistic function and $P$ the diagonal prior precision
matrix. The residual rank and prior precision are source-selected.
Appendix~\ref{sec:implementation} specifies initialization, scaling, and
the configuration grid.

\paragraph{Source response signatures.}
We define source-signature groups by exact equality of source responses:
$i\sim j$ if $M_{si}=M_{sj}$ for every source agent $s$.
Each group $\mathcal G_g$ contains $N_g$ tasks. Tasks sharing a signature
are observationally indistinguishable in the source matrix, so we
average their source-fitted intercepts, slopes, and residual loadings.
With $\ell_g=(a_g,v_g^\top)^\top$ collecting the averaged loadings,
the group baseline success probability is
\begin{equation}
 p_g(z)=\sigma(b_g+\ell_g^\top z).
\end{equation}
We model target outcomes within each group as exchangeable draws from
a shared uncertain success rate.
Unobserved outcomes within a group need not be identical.

\subsection{Target-Specific Group Uncertainty}
The group success rate $q_g$ has conditional prior mean $p_g(z)$:
\begin{align}
 q_g\mid z&\sim\operatorname{Beta}
       \big(c_g(z)p_g(z),c_g(z)(1-p_g(z))\big),\\
 c_g(z)&=\frac{\kappa}{p_g(z)(1-p_g(z))},\qquad \kappa>0.
 \label{eq:concentration}
\end{align}
Group rates are independent conditional on $z$.
Given these rates, outcomes $Y_i$ for $i\in\mathcal G_g$ are
independent Bernoulli variables with rate $q_g$. Integrating $q_g$ couples tasks within a group
even after conditioning on the global state: they share an unknown
target-specific rate. For a fixed $p_g(z)$, smaller $\kappa$ allows greater
prior dispersion around this mean. A single source-selected $\kappa$
applies to all groups. The probability-dependent concentration gives
smaller prior variance when the baseline probability is near zero or one.

Let $n_g$ and $s_g$ count observed outcomes and successes in group $g$.
Conditional conjugacy gives
\begin{equation}
 q_g\mid z,D_t\sim\operatorname{Beta}(A_g,F_g),
 \label{eq:conditional}
\end{equation}
where $A_g=c_gp_g+s_g$ and $F_g=c_g(1-p_g)+n_g-s_g$.
The observed history also updates the target latent-state posterior:
\begin{equation}
 p(z\mid D_t)\propto p(z)
 \prod_g\frac{\mathrm B(A_g,F_g)}
 {\mathrm B(c_gp_g,c_g(1-p_g))}.
 \label{eq:latentposterior}
\end{equation}
Here $\mathrm B$ is the Beta function; the ratio gives the likelihood of
the observed binary sequence within each group. An observation updates
the group-rate and latent-state posteriors, sharing information within
and across groups, respectively.

With $U_g=N_g-n_g$ unqueried tasks in group $g$, the unobserved success
count $K_g=\sum_{i\in\mathcal G_g\setminus C_t}Y_i$ has distribution
\begin{equation}
 K_g\mid z,D_t\sim\operatorname{BetaBinomial}(U_g,A_g,F_g).
 \label{eq:predictive}
\end{equation}
The finite score satisfies $NS=\sum_gs_g+\sum_g K_g$. Conditional on $z$
the group counts are independent; marginally they share uncertainty
about $z$. The predictive score distribution retains these dependencies
through joint integration over the global state and group rates.

\subsection{Posterior-Guided Acquisition}
\label{sec:acquisition}
For a group with $U_g>0$, let $Y_g$ denote the outcome of a task sampled
uniformly from $\mathcal G_g\setminus C_t$.
The score-variance acquisition rule selects the group with maximum
\begin{equation}
 G_g^{S}=
 \frac{\operatorname{Cov}(S,Y_g\mid D_t)^2}
 {\operatorname{Var}(Y_g\mid D_t)}.
 \label{eq:gain}
\end{equation}
\begin{proposition}[Score variance reduction]
\label{prop:score-gain}
Under the HSPR predictive model, for any unobserved binary response $Y_g$
with $0<\Pr(Y_g=1\mid D_t)<1$, Equation~\ref{eq:gain} equals the
expected reduction in posterior variance of the full-benchmark score:
\begin{align}
 G_g^S&=\operatorname{Var}(S\mid D_t)\notag\\
 &\quad-\E\big[\operatorname{Var}(S\mid D_t,Y_g)\mid D_t\big].
 \label{eq:gain-identity}
\end{align}
\end{proposition}
This greedy criterion includes resolving the queried outcome and updating
predictions for other unqueried tasks. The variance identity follows from
the law of total variance (Appendix~\ref{sec:gain-proof}); it does not
establish optimality for final absolute error.

The covariance shows how hierarchy enters task selection.
Let $m_g=\E[q_g\mid z,D_t]$ and
$\tau_g^2=\operatorname{Var}(q_g\mid z,D_t)$. Conditional on the global state,
\begin{align}
 \operatorname{Cov}(S,Y_g\mid z,D_t)
 &=\frac{m_g(1-m_g)}{N}\notag\\
 &\quad+\frac{(U_g-1)\tau_g^2}{N}.
 \label{eq:group-sharing}
\end{align}
The terms represent resolving the queried outcome and sharing evidence
with the other $U_g-1$ group members. Larger unqueried groups amplify
the latter contribution at a fixed rate posterior. Integrating over $z$
adds dependencies across groups.

The latent-state acquisition alternative uses
\begin{equation}
 G_g^{z}=\sum_k
 \frac{\operatorname{Cov}(z_k,Y_g\mid D_t)^2}
 {\operatorname{Var}(Y_g\mid D_t)}.
 \label{eq:latentgain}
\end{equation}
It equals the expected reduction in the sum of posterior variances of
the latent coordinates under the same model.
Source-only validation chooses the acquisition objective before target
replay: 13 of the 16 deployed folds select Score and three select
Latent. Both use the same hierarchical posterior; the final
absolute-loss decision is specified separately below.

After choosing a group, we sample uniformly from its unqueried tasks.
Exhausted groups are excluded. The selected task is evaluated once,
$(n_g,s_g)$ are updated, and the posterior is recomputed before the next
query. Uniform within-group selection treats tasks with the same source
signature symmetrically.

\subsection{Predictive Score Decision}
Under absolute score loss, a median $S_{.5}$ of $p(S\mid D_B)$ minimizes
posterior expected loss:
\begin{equation}
 S_{.5}\in\arg\min_x\E[|S-x|\mid D_B].
 \label{eq:median}
\end{equation}
We sample $z$ from its weighted approximation, then each conditional
$q_g$, and finally $K_g\sim\operatorname{Binomial}(U_g,q_g)$.
Every score draw includes the observed successes exactly. The median of these
complete score draws estimates $S_{.5}$.

For optional calibration, let
$\bar q_g=\E[q_g\mid D_B]$ and define
\begin{align}
 R_B&=\frac{\sum_gs_g-\sum_gn_g\bar q_g}{B},\\
 \widehat S&=\operatorname{clip}_{[0,1]}
 \left[S_{.5}+\lambda(1-B/N)R_B\right].
 \label{eq:calibration}
\end{align}
Source-only validation selects $\lambda\in\{0,.5\}$ for the median decision.
The factor $(1-B/N)$ limits correction to the unobserved fraction.
This plug-in calibration uses the same observed outcomes as posterior fitting.
The absolute-loss optimality in Equation~\ref{eq:median} applies to the
unadjusted predictive median. At full budget, $\widehat S=S$.

\subsection{Source-Only Configuration}
Source-only leave-one-agent-out validation proceeds in stages.
The global stage jointly evaluates rank, precision, acquisition, and
calibration, retaining the selected rank and precision. With these fixed,
the hierarchical stage selects group uncertainty strength and acquisition
jointly with posterior-mean calibration. The final stage selects median
calibration on the resulting query paths. Each validation row is removed
before fitting item parameters, residual factors, and groups. We refit
on all source agents and freeze the configuration before target evaluation.
Appendix~\ref{sec:implementation} gives the grids and fitting details.

We approximate the continuous posterior in Equation~\ref{eq:latentposterior}
by importance sampling with a Gaussian proposal centered at an auxiliary
logistic-normal MAP estimate. The weights account for the target-model
likelihood, prior, and proposal density. Predictive draws estimate the
finite-score median; Appendix~\ref{sec:numerical-checks} reports numerical
validation.

\begin{algorithm}[t]
\caption{HSPR evaluation of a target agent}
\label{alg:hspr}
\small
\begin{algorithmic}[1]
\Require Source responses $M$, target execution oracle $\mathcal O$, budget $B$
\State Select $(r,P,\kappa,\rho,\lambda)$ by staged source-only validation, $\rho\in\{S,z\}$
\State Fit item parameters and source-signature groups on $M$
\State Initialize joint posterior; $D_0,C_0\gets\emptyset$
\For{$t=0,\ldots,B-1$}
    \State $g\gets\arg\max_{h:U_h>0}G_h^\rho(D_t)$
    \State $i\sim\operatorname{Uniform}(\mathcal G_g\setminus C_t)$
    \State $D_{t+1}\gets D_t\cup\{(i,\mathcal O(i))\}$
    \State $C_{t+1}\gets C_t\cup\{i\}$
    \State Update global and conditional group posteriors
\EndFor
\State Construct $p(S\mid D_B)$ from observed successes and predictive unobserved success counts
\State \Return Calibrated score estimate $\widehat S$ (Eq.~\ref{eq:calibration})
\end{algorithmic}
\end{algorithm}

\section{Experiments}
\label{sec:results}

\begin{table*}[t]
\centering\scriptsize
\setlength{\tabcolsep}{2.8pt}
\renewcommand{\arraystretch}{1.15}
\resizebox{\textwidth}{!}{%
\begin{tabular}{@{}l*{12}{r}@{}}\toprule
Method & \multicolumn{3}{c}{Lite (30/300)} & \multicolumn{3}{c}{Verified (50/500)} & \multicolumn{3}{c}{Full (229/2294)} & \multicolumn{3}{c}{Pro (73/730)} \\
\cmidrule(lr){2-4}\cmidrule(lr){5-7}\cmidrule(lr){8-10}\cmidrule(lr){11-13}
 & MAE $\downarrow$ & MaxErr $\downarrow$ & $\tau_b$ $\uparrow$ & MAE $\downarrow$ & MaxErr $\downarrow$ & $\tau_b$ $\uparrow$ & MAE $\downarrow$ & MaxErr $\downarrow$ & $\tau_b$ $\uparrow$ & MAE $\downarrow$ & MaxErr $\downarrow$ & $\tau_b$ $\uparrow$ \\
\midrule
Uniform raw & 6.632 & 19.300 & 0.654 & 5.482 & 15.460 & 0.785 & 2.227 & 5.342 & 0.923 & 3.832 & 10.233 & 0.757 \\
Stratified & 4.468 & 13.354 & 0.723 & 3.302 & 10.768 & 0.843 & 1.636 & 3.877 & 0.938 & 3.111 & \underline{7.508} & 0.837 \\
RSL \citep{zhang2025benchmark} & 4.911 & 16.064 & 0.683 & 3.654 & 14.590 & 0.823 & 2.277 & 8.547 & 0.930 & 4.467 & 12.822 & 0.714 \\
FAQ-port \citep{wu2026faq} & 6.681 & 26.263 & 0.638 & 4.951 & 25.989 & 0.765 & 1.807 & 5.049 & 0.921 & 5.433 & 15.330 & 0.673 \\
Mid-Range \citep{ndzomga2026efficient} & 18.448 & 49.333 & 0.830 & 16.449 & 32.600 & \textbf{0.915} & 24.709 & 50.239 & 0.862 & 22.798 & 39.041 & \textbf{0.967} \\
Mid-Range + Ridge (control) & 4.717 & 19.223 & 0.725 & 3.653 & 18.734 & 0.844 & 1.532 & 10.394 & 0.890 & 3.434 & 10.980 & 0.868 \\
ATLAS-port \citep{li2026atlas} & \underline{2.689} & \underline{8.989} & \underline{0.851} & \textbf{1.861} & \underline{7.406} & 0.914 & \underline{0.957} & \underline{2.945} & \textbf{1.000} & 7.736 & 14.855 & 0.398 \\
PTA-IRT \citep{duan2026pta} & 2.724 & 11.703 & 0.816 & 2.177 & 8.310 & 0.887 & 2.280 & 6.174 & 0.956 & \underline{2.731} & 11.729 & 0.890 \\
\midrule
\textbf{\method{} (ours)} & \textbf{1.927} & \textbf{5.683} & \textbf{0.912} & \underline{1.915} & \textbf{6.000} & \underline{0.914} & \textbf{0.717} & \textbf{2.638} & \underline{0.978} & \textbf{2.098} & \textbf{4.601} & \underline{0.895} \\
\bottomrule\end{tabular}}
\caption{Main results at a matched 10\% task budget. MAE and MaxErr are in percentage points (pp); $\tau_b$ is unscaled. Metrics pool complete target cohorts before averaging query seeds. Best and second-best point estimates are bold and underlined, respectively, using unrounded values. Section~\ref{sec:protocol} describes the baselines and ports.}
\label{tab:low-cost-leaderboard}
\end{table*}

\subsection{Experimental Setup}
\label{sec:protocol}
We evaluate held-out agent configurations on four archived task banks
from the SWE-bench family. We use public PTA-IRT response matrices
\citep{duan2026pta} for Lite, Verified, Full \citep{jimenez2024swebench},
and Pro \citep{deng2026swebenchpro}. The four task banks contain 300, 500,
2,294, and 730 tasks and 35, 70, 14, and 14 agent configurations,
respectively. A configuration is an evaluated agent setup, represented
by one row of the response matrix. Each benchmark uses four folds. Source and target
configuration identifiers are disjoint within a fold; all configurations
are targets once. This configuration-level split permits shared base
models or related scaffolds between sources and targets.

Source folds contain 26--27 configurations on Lite, 52--53 on Verified,
and 10--11 on Full and Pro. At the primary 10\% checkpoint, budgets are
$B=\operatorname{round}(.1N)$: 30, 50, 229, and 73 distinct tasks.
HSPR model settings, acquisition objectives, and calibration are frozen
after source-only validation and before target replay.
Each target has five HSPR query paths (seeds 0--4), each containing $B$
distinct tasks. Budgets apply separately to each method and query path.
Predictions are recorded before full target outcomes are accessed for scoring.
The historical response bank is treated as given, and $B$ counts
distinct target-task outcomes revealed during replay. This budget
represents task executions, without rerunning the archived agents.

We compare Uniform raw means, difficulty stratification,
RSL \citep{zhang2025benchmark}, FAQ-port \citep{wu2026faq},
Mid-Range raw means \citep{ndzomga2026efficient}, Mid-Range with Ridge,
ATLAS-port \citep{li2026atlas}, and frozen PTA-IRT \citep{duan2026pta}.
RSL applies source-fitted RidgeCV with penalties $(.1,1,10)$ to uniformly
sampled tasks, sharing its query sets with Uniform raw.
Mid-Range selects the $B$ tasks whose source pass rates are closest to .5.
Mid-Range with Ridge shares the Mid-Range raw query set: a source-fitted
regression maps the selected outcomes to full scores.
FAQ-port uses without-replacement queries and caps factor rank for the
available source cohort. ATLAS-port retains Fisher-guided selection and
p-IRT estimation, using a source-fold 4PL item bank. PTA-IRT uses frozen
released predictions on the same configurations.
Appendix~\ref{sec:accounting} details these adaptations.

Table~\ref{tab:low-cost-leaderboard} compares complete procedures under
matched target-task budgets, allowing their query policies and estimators
to differ. Table~\ref{tab:hspr-controls} examines group uncertainty,
adaptive querying, and the predictive decision under controlled settings.

\paragraph{Metrics.}
For each task bank, let $\mathcal A$ denote the set of held-out agent
configurations pooled across all four folds. For each method and query seed, we compute
\begin{align}
 \mathrm{MAE}&=\frac{1}{|\mathcal A|}\sum_{a\in\mathcal A}
                      |\widehat S_a-S_a|,\\
 \mathrm{MaxErr}&=\max_{a\in\mathcal A}|\widehat S_a-S_a|.
\end{align}
Kendall's $\tau_b$ uses all cohort pairs and accounts for ties.
We then average the pooled metrics over query seeds. Sampling/RSL
baselines have ten saved seeds, ATLAS-port and HSPR five, and deterministic
or frozen methods one prediction per configuration. MAE and MaxErr are
reported in pp throughout.

Appendix~\ref{sec:compute-overhead} reports HSPR evaluator overhead.


\begin{table*}[t]\centering\small
\begin{tabular}{lrrrrr}\toprule
Ablation / control & Lite $\downarrow$ & Verified $\downarrow$ & Full $\downarrow$ & Pro $\downarrow$ & Macro $\downarrow$ \\ \midrule
\textbf{\method{} (ours)} & 1.927 & 1.915 & 0.717 & 2.098 & 1.664 \\
\midrule\multicolumn{6}{l}{\textit{A. Hierarchical modeling: source-selected acquisition objective}} \\
w/o group uncertainty & 1.895 & 1.970 & 1.331 & 2.034 & 1.808 \\
\midrule\multicolumn{6}{l}{\textit{B. Query adaptivity: frozen HSPR reconstruction}} \\
w/o adaptive queries (Uniform) & 4.573 & 3.288 & 1.413 & 3.066 & 3.085 \\
w/o adaptive queries (Mid-Range) & 3.961 & 3.059 & 2.741 & 1.975 & 2.934 \\
\midrule\multicolumn{6}{l}{\textit{C. Predictive decision: identical HSPR query paths}} \\
w/o median (posterior mean) & 1.941 & 1.927 & 0.722 & 2.108 & 1.674 \\
\bottomrule\end{tabular}
\caption{Ablations at a 10\% budget; MAE in pp, with equal-bank Macro. A retains each fold's acquisition objective, global configuration, and median calibration; only group uncertainty changes, and both variants generate their own query paths. B freezes HSPR reconstruction and calibration, using five Uniform seeds or one deterministic Mid-Range path. C shares HSPR query paths, with decision-specific calibration selected on sources.}\label{tab:hspr-controls}
\end{table*}


\begin{table*}[t]\centering\small
\begin{tabular}{lrrrrr}\toprule
Bank & Tasks $N$ & Groups $\bar G$ & $N/\bar G$ & \multicolumn{2}{c}{Source-zero tasks} \\
\cmidrule(lr){5-6}
 & & & & Share (\%) & Target success (\%) \\ \midrule
Lite & 300 & 201.5 & 1.49 & 24.6 & 0.87 \\
Verified & 500 & 426.0 & 1.17 & 8.3 & 0.74 \\
Full & 2294 & 275.9 & \textbf{8.32} & 41.3 & 1.84 \\
Pro & 730 & 222.0 & 3.29 & 36.7 & 2.07 \\
\bottomrule\end{tabular}
\caption{Historical response structure. $\bar G$ averages source group counts over targets; $N/\bar G$ divides the task count by this average. Source-zero tasks have no successes among the source agents. Target success rates within these sets are averaged over targets, using complete outcomes after prediction.}\label{tab:hspr-mechanism}
\end{table*}

\begin{figure*}[t]
\centering
\includegraphics[width=\textwidth]{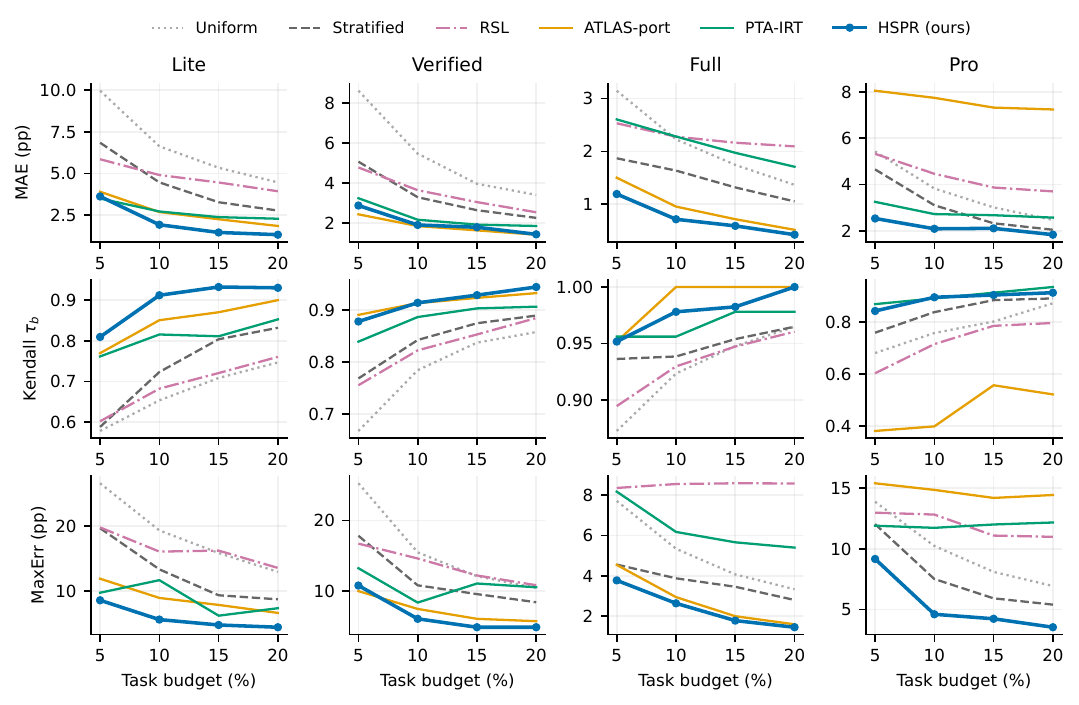}
\caption{Budget sensitivity across four SWE-bench task banks at 5--20\% task budgets.
HSPR uses fixed 10\% source-selected settings and nested query paths;
baselines retain budget-specific settings. MAE and MaxErr are in pp;
metrics pool target configurations before averaging query seeds.}
\label{fig:budget-curves}
\end{figure*}

\subsection{Main Results}
At a 10\% task budget, HSPR achieves the lowest MAE point estimates on Lite, Full, and
Pro and remains within 0.055 pp of ATLAS-port on Verified
(Table~\ref{tab:low-cost-leaderboard}). It has the lowest seed-averaged MaxErr on all four banks and ranks
among the top two in all 12 benchmark--metric combinations, leading
eight. Equal-bank macro MAE is 1.664 pp, compared with 2.478 pp for
PTA-IRT, the strongest baseline on this aggregate.

On Pro, raw Mid-Range preserves ordering ($\tau_b=.967$) but has
MAE 22.798 pp versus 2.098 pp for HSPR. This difference motivates
reporting absolute score error alongside ranking fidelity.

Holm-corrected Wilcoxon tests on paired, seed-averaged absolute errors
favor HSPR over RSL on all four banks, over Stratified on three, and
over ATLAS-port on Pro.
Appendix~\ref{sec:paired-comparisons} reports the complete bootstrap
intervals and corrected tests.

\subsection{Ablation Study}
\paragraph{Hierarchical modeling.}
Hierarchical uncertainty affects the evaluation procedure beyond its
final reconstruction step. Table~\ref{tab:hspr-controls}A holds the
source-selected acquisition objective fixed within each fold and retains the adopted global
configuration and median calibration. The two variants generate their
own query paths, so removing group uncertainty changes both the evidence
collected and its interpretation. Removing hierarchy increases equal-bank
macro MAE from 1.664 to 1.808 pp; the paired configuration bootstrap
interval for the 0.143 pp difference is $[-0.012,0.302]$.
The largest gain occurs on Full, where hierarchy reduces MAE from
1.331 to 0.717 pp. Appendix~\ref{sec:numerical-checks}
reports the execution protocol and a separate reconstruction control
holding query paths fixed.

\paragraph{Query adaptivity.}
Informative query allocation provides a large aggregate gain with the
reconstructor held fixed. Table~\ref{tab:hspr-controls}B freezes the
HSPR model, predictive median, and calibration and replaces adaptive
queries with Uniform or fixed Mid-Range queries. Macro MAE rises from
1.664 to 3.085 and 2.934 pp, respectively. HSPR updates acquisition scores
after each observed outcome, allowing subsequent queries to reflect the
current target posterior.

\paragraph{Predictive decision.}
Table~\ref{tab:hspr-controls}C uses identical HSPR query paths and
replaces the median with the posterior mean, with calibration selected
on sources for each decision family. The median-based estimate has
slightly lower macro MAE (1.664 versus 1.674 pp).

\subsection{Mechanism Analysis}
\label{sec:hspr-mechanism}
Source-signature groups support local evidence sharing.
Full's 2,294 tasks form approximately 276 source-signature groups on
average across targets, giving $N/\bar G=8.32$
(Table~\ref{tab:hspr-mechanism}). An observed outcome updates the posterior
over the shared group success rate, informing predictions for unqueried
group members.
Verified retains 426 groups among 500 tasks, leaving less within-group
sharing to exploit. This structural contrast is consistent with the
larger end-to-end hierarchical gain on Full. Larger groups allow more
within-group pooling; its usefulness also depends on how the
historical structure transfers to the target.

Source-zero tasks are those with no successes among the agents in a
given source fold. On Full and Pro,
41.3\% and 36.7\% of tasks have all-zero source responses, yet target
success rates within these sets are 1.84\% and 2.07\%, respectively.
These post-hoc target rates use complete outcomes after estimation.
The tasks share a source signature within a fold while admitting
different target outcomes. An uncertain target-specific rate represents
this departure and lets observed evidence revise predictions for
remaining group members.

Hierarchy also changes where evidence is acquired. In the controlled
Full comparison, source-zero tasks receive 13.2\% of queries with
hierarchy, compared with 0.19\% without it.

\subsection{Budget Sensitivity}
Across 5--20\% task budgets, HSPR's equal-bank macro MAE decreases from
2.562 to 1.262 pp (Figure~\ref{fig:budget-curves}). Its 10\% source-selected settings remain
fixed at all checkpoints, with nested distinct-task paths and five query
seeds per target. Baselines retain their saved budget-specific settings.

\subsection{Discussion}
Conditional on the target latent state, the second term in
Equation~\ref{eq:group-sharing} scales with the number of other unqueried
group members and the posterior variance of their common rate.
At a fixed task-bank size and rate posterior, an observation can inform
more remaining outcomes in a larger group. Full's larger source-signature
groups (Table~\ref{tab:hspr-mechanism}) permit more within-group sharing
than Verified's, consistent with its larger hierarchical gain
(Table~\ref{tab:hspr-controls}). As queries reveal outcomes, the remaining
group sizes and rate posteriors change, altering this contribution to
Score acquisition.

\section{Conclusion}
HSPR combines source-fitted response factors, uncertain group success
rates, and posterior-guided acquisition to estimate full-benchmark scores
from limited observations. Its final estimate uses a posterior median
with source-selected calibration. At a 10\% task budget, HSPR achieves the
lowest MAE point estimates on three of four SWE-bench task banks and the
lowest MaxErr on all four among the compared methods. Controlled experiments
show substantial aggregate gains from adaptive querying. The contribution
of group uncertainty varies across task banks, with the largest observed
improvement on SWE-bench Full. These results highlight the role of
historical response structure in task-efficient score recovery.

\section*{Limitations}
We evaluate held-out agent configurations on four task banks from the
SWE-bench family. Generalization under substantial shifts in model families
or agent scaffolds remains to be evaluated. HSPR's response factors and
source-signature groups are derived from the historical source bank and
therefore depend on its coverage. Posterior uncertainty is conditional on
the assumed response model, including within-group exchangeability.

We measure evaluation budgets by the number of distinct target-task
executions and report evaluator inference overhead in
Appendix~\ref{sec:compute-overhead}. Token, wall-clock, and monetary savings
depend on per-task execution costs and cannot be inferred directly from
task-count reduction. Historical data collection, source fitting, and
configuration selection incur additional costs.

\section*{Ethical Considerations}
This study uses existing software-agent response matrices and does not
involve human participants or private user data. Treating recovered scores
as fully observed results may mislead model comparisons, particularly when
historical source data poorly represent a target configuration. We therefore
recommend reporting the evaluation budget and clearly distinguishing
recovered scores from full-benchmark measurements.

\bibliography{references}

\begin{thebibliography}{27}
\providecommand{\natexlab}[1]{#1}

\bibitem[{Ayyad et~al.(2026)Ayyad, Wang, Shin, Zou, and Adams}]{ayyad2026sail}
Mahmoud Ayyad, Zehao Wang, Jiho Shin, Ying Zou, and Bram Adams. 2026.
\newblock \href {https://arxiv.org/abs/2609.24928} {Trajectory-aware benchmark
  subset selection for cost-efficient software engineering agent regression
  testing}.
\newblock \emph{arXiv preprint arXiv:2609.24928}.

\bibitem[{Chen et~al.(2026)Chen, Yang, Yang, Sha, Li, Wang, Xian, Kong, Liu,
  and Bian}]{chen2026agent2rlbench}
Wanyi Chen, Xiao Yang, Xu~Yang, Tianming Sha, Qizheng Li, Zhuo Wang, Bowen
  Xian, Fang Kong, Weiqing Liu, and Jiang Bian. 2026.
\newblock \href {https://doi.org/10.48550/arXiv.2604.10547} {{Agent}$^2$
  {RL-Bench}: Can {LLM} agents engineer agentic {RL} post-training?}
\newblock \emph{arXiv preprint arXiv:2604.10547}.

\bibitem[{Deng et~al.(2026)Deng, Da, Pan, He, Ide, Garg, Lauffer, Park, Rane,
  Sampath, Krishnan, Kundurthy, Hendryx, Wang, Zhang, Jacobson, Liu, and
  Kenstler}]{deng2026swebenchpro}
Xiang Deng, Jeff Da, Edwin Pan, Yannis~Yiming He, Charles Ide, Kanak Garg,
  Niklas Lauffer, Andrew Park, Chetan Rane, Karmini Sampath, Maya Krishnan,
  Srivatsa~R Kundurthy, Sean~M. Hendryx, Zifan Wang, Chen Bo~Calvin Zhang, Noah
  Jacobson, Bing Liu, and Brad Kenstler. 2026.
\newblock \href {https://proceedings.mlr.press/v306/deng26f.html} {{SWE}-bench
  pro: Can {AI} agents solve long-horizon software engineering tasks?}
\newblock In \emph{Proceedings of the 43rd International Conference on Machine
  Learning}, pages 23917--23934.

\bibitem[{Duan et~al.(2026)Duan, Zheng, Wang, Wang, Shi, Liu, Ma, Chen, Liu,
  and Zheng}]{duan2026pta}
Kefeng Duan, Dewu Zheng, Yanlin Wang, Xiwen Wang, Ensheng Shi, Xilin Liu, Yuchi
  Ma, Jiachi Chen, Mingwei Liu, and Zibin Zheng. 2026.
\newblock \href {https://arxiv.org/abs/2609.01603} {Efficient {SWE} agent
  benchmarking via trajectory-aware evaluation}.
\newblock \emph{arXiv preprint arXiv:2609.01603}.

\bibitem[{Ge et~al.(2026)Ge, Kryvosheieva, Fried, Girit, and
  Hariharan}]{ge2026psychometrics}
Chris Ge, Daria Kryvosheieva, Daniel Fried, Uzay Girit, and Kaivalya Hariharan.
  2026.
\newblock \href {https://arxiv.org/abs/2604.00594} {Agent psychometrics:
  Task-level performance prediction in agentic coding benchmarks}.
\newblock In \emph{Proceedings of the Conference on Language Modeling
  ({COLM})}.

\bibitem[{Gusev and Zaytsev(2026)}]{gusev2026ranking}
Rostislav Gusev and Alexey Zaytsev. 2026.
\newblock \href {https://doi.org/10.1145/3770855.3817569} {Benchmarking on
  tasks that matter: Dataset selection for preserving model rankings}.
\newblock In \emph{Proceedings of the 32nd ACM SIGKDD Conference on Knowledge
  Discovery and Data Mining V.2}, pages 8954--8965.

\bibitem[{Hendrycks et~al.(2021)Hendrycks, Burns, Basart, Zou, Mazeika, Song,
  and Steinhardt}]{hendrycks2021mmlu}
Dan Hendrycks, Collin Burns, Steven Basart, Andy Zou, Mantas Mazeika, Dawn
  Song, and Jacob Steinhardt. 2021.
\newblock \href {https://openreview.net/forum?id=d7KBjmI3GmQ} {Measuring
  massive multitask language understanding}.
\newblock In \emph{International Conference on Learning Representations}.

\bibitem[{Jimenez et~al.(2024)Jimenez, Yang, Wettig, Yao, Pei, Press, and
  Narasimhan}]{jimenez2024swebench}
Carlos~E. Jimenez, John Yang, Alexander Wettig, Shunyu Yao, Kexin Pei, Ofir
  Press, and Karthik Narasimhan. 2024.
\newblock \href {https://openreview.net/forum?id=VTF8yNQM66} {{SWE-bench}: Can
  language models resolve real-world {GitHub} issues?}
\newblock In \emph{International Conference on Learning Representations}.

\bibitem[{Kiela et~al.(2021)Kiela, Bartolo, Nie, Kaushik, Geiger, Wu
  et~al.}]{kiela2021dynabench}
Douwe Kiela, Max Bartolo, Yixin Nie, Divyansh Kaushik, Atticus Geiger,
  Zhengxuan Wu, et~al. 2021.
\newblock \href {https://doi.org/10.18653/v1/2021.naacl-main.324} {{Dynabench}:
  Rethinking benchmarking in {NLP}}.
\newblock In \emph{Proceedings of the 2021 Conference of the North American
  Chapter of the Association for Computational Linguistics: Human Language
  Technologies}.

\bibitem[{Kossen et~al.(2022)Kossen, Farquhar, Gal, and
  Rainforth}]{kossen2022surrogate}
Jannik Kossen, Sebastian Farquhar, Yarin Gal, and Thomas Rainforth. 2022.
\newblock \href {https://doi.org/10.52202/068431-1783} {Active surrogate
  estimators: An active learning approach to label-efficient model evaluation}.
\newblock In \emph{Advances in Neural Information Processing Systems},
  volume~35.

\bibitem[{Kossen et~al.(2021)Kossen, Farquhar, Gal, and
  Rainforth}]{kossen2021active}
Jannik Kossen, Sebastian Farquhar, Yarin Gal, and Tom Rainforth. 2021.
\newblock \href {https://proceedings.mlr.press/v139/kossen21a.html} {Active
  testing: Sample-efficient model evaluation}.
\newblock In \emph{Proceedings of the 38th International Conference on Machine
  Learning}, pages 5753--5763.

\bibitem[{Lalor et~al.(2016)Lalor, Wu, and Yu}]{lalor2016scale}
John~P. Lalor, Hao Wu, and Hong Yu. 2016.
\newblock \href {https://doi.org/10.18653/v1/D16-1062} {Building an evaluation
  scale using item response theory}.
\newblock In \emph{Proceedings of the 2016 Conference on Empirical Methods in
  Natural Language Processing}, pages 648--657.

\bibitem[{Li et~al.(2026)Li, Tang, Chen, Cheng, Metoyer, Hua, and
  Chawla}]{li2026atlas}
Peiyu Li, Xiuxiu Tang, Si~Chen, Ying Cheng, Ronald Metoyer, Ting Hua, and
  Nitesh~V. Chawla. 2026.
\newblock \href {https://proceedings.mlr.press/v306/li26fx.html} {Adaptive
  testing for {LLM} evaluation: A psychometric alternative to static
  benchmarks}.
\newblock In \emph{Proceedings of the 43rd International Conference on Machine
  Learning}, pages 70246--70269.

\bibitem[{Liang et~al.(2023)Liang, Bommasani, Lee, Tsipras, Soylu, Yasunaga,
  Zhang, Narayanan et~al.}]{liang2022helm}
Percy Liang, Rishi Bommasani, Tony Lee, Dimitris Tsipras, Dilara Soylu,
  Michihiro Yasunaga, Yian Zhang, Deepak Narayanan, et~al. 2023.
\newblock \href {https://openreview.net/forum?id=iO4LZibEqW} {Holistic
  evaluation of language models}.
\newblock \emph{Transactions on Machine Learning Research}.

\bibitem[{Liu et~al.(2024)Liu, Yu, Zhang, Xu, Lei, Lai
  et~al.}]{liu2023agentbench}
Xiao Liu, Hao Yu, Hanchen Zhang, Yifan Xu, Xuanyu Lei, Hanyu Lai, et~al. 2024.
\newblock \href {https://openreview.net/forum?id=zAdUB0aCTQ} {{AgentBench}:
  Evaluating {LLM}s as agents}.
\newblock In \emph{International Conference on Learning Representations}.

\bibitem[{Liu et~al.(2025)Liu, Li, Zhuang, Liu, Shen, Ouyang, Cheng, and
  Wang}]{liu2025amelo}
Zirui Liu, Jiatong Li, Yan Zhuang, Qi~Liu, Shuanghong Shen, Jie Ouyang, Mingyue
  Cheng, and Shijin Wang. 2025.
\newblock \href {https://proceedings.mlr.press/v267/liu25ak.html} {am-{ELO}: A
  stable framework for arena-based {LLM} evaluation}.
\newblock In \emph{Proceedings of the 42nd International Conference on Machine
  Learning}, pages 38857--38868.

\bibitem[{Ma et~al.(2024)Ma, Zhang, Zhu, Yang, Yang, Jin, Lan, Kong, and
  He}]{ma2024agentboard}
Chang Ma, Junlei Zhang, Zhihao Zhu, Cheng Yang, Yujiu Yang, Yaohui Jin,
  Zhenzhong Lan, Lingpeng Kong, and Junxian He. 2024.
\newblock \href {https://doi.org/10.52202/079017-2365} {{AgentBoard}: An
  analytical evaluation board of multi-turn {LLM} agents}.
\newblock In \emph{Advances in Neural Information Processing Systems},
  volume~37, pages 74325--74362.

\bibitem[{Maia~Polo et~al.(2024)Maia~Polo, Weber, Choshen, Sun, Xu, and
  Yurochkin}]{polo2024tiny}
Felipe Maia~Polo, Lucas Weber, Leshem Choshen, Yuekai Sun, Gongjun Xu, and
  Mikhail Yurochkin. 2024.
\newblock \href {https://proceedings.mlr.press/v235/maia-polo24a.html}
  {tiny{B}enchmarks: Evaluating {LLM}s with fewer examples}.
\newblock In \emph{Proceedings of the 41st International Conference on Machine
  Learning}, pages 34303--34326.

\bibitem[{Ndzomga(2026)}]{ndzomga2026efficient}
Franck Ndzomga. 2026.
\newblock \href {https://doi.org/10.48550/arXiv.2603.23749} {Efficient
  benchmarking of {AI} agents}.
\newblock \emph{arXiv preprint arXiv:2603.23749}.

\bibitem[{Saranathan et~al.(2025)Saranathan, Xu, Alam, Kumar, Foltin, Wong, and
  Bhattacharya}]{saranathan2025sublime}
Gayathri Saranathan, Cong Xu, Mahammad~Parwez Alam, Tarun Kumar, Martin Foltin,
  Soon~Yee Wong, and Suparna Bhattacharya. 2025.
\newblock \href {https://doi.org/10.18653/v1/2025.acl-long.1477} {{SubLIME}:
  Subset selection via rank correlation prediction for data-efficient {LLM}
  evaluation}.
\newblock In \emph{Proceedings of the 63rd Annual Meeting of the Association
  for Computational Linguistics (Volume 1: Long Papers)}, pages 30572--30593.

\bibitem[{Shi et~al.(2026)Shi, Sun, Dong, Wan, Lo, and Gu}]{shi2026earlyeval}
Yuling Shi, Zhensu Sun, Junsen Dong, Chengcheng Wan, David Lo, and Xiaodong Gu.
  2026.
\newblock \href {https://arxiv.org/abs/2609.02783} {{EarlyEval}: Cheaper agent
  evaluation via early outcome prediction}.
\newblock \emph{arXiv preprint arXiv:2609.02783}.

\bibitem[{Srivastava et~al.(2023)Srivastava, Rastogi, Rao, Shoeb, Abid, Fisch,
  Brown et~al.}]{srivastava2022beyond}
Aarohi Srivastava, Abhinav Rastogi, Abhishek Rao, Abu Awal~Md Shoeb, Abubakar
  Abid, Adam Fisch, Adam~R. Brown, et~al. 2023.
\newblock \href {https://openreview.net/forum?id=uyTL5Bvosj} {Beyond the
  imitation game: Quantifying and extrapolating the capabilities of language
  models}.
\newblock \emph{Transactions on Machine Learning Research}.

\bibitem[{Truong et~al.(2025)Truong, Tu, Liang, Li, and
  Koyejo}]{truong2025amortized}
Sang~T. Truong, Yuheng Tu, Percy Liang, Bo~Li, and Sanmi Koyejo. 2025.
\newblock \href {https://proceedings.mlr.press/v267/truong25c.html} {Reliable
  and efficient amortized model-based evaluation}.
\newblock In \emph{Proceedings of the 42nd International Conference on Machine
  Learning}, pages 60238--60265.

\bibitem[{van~der Linden(2008)}]{vanderlinden2008response}
Wim~J. van~der Linden. 2008.
\newblock \href {https://doi.org/10.3102/1076998607302626} {Using response
  times for item selection in adaptive testing}.
\newblock \emph{Journal of Educational and Behavioral Statistics}, 33(1):5--20.

\bibitem[{Wu et~al.(2026)Wu, Nair, and Cand\`es}]{wu2026faq}
Skyler Wu, Yash Nair, and Emmanuel~J. Cand\`es. 2026.
\newblock \href {https://arxiv.org/abs/2601.20251} {Efficient evaluation of
  {LLM} performance with statistical guarantees}.
\newblock \emph{arXiv preprint arXiv:2601.20251}.

\bibitem[{Zhang et~al.(2025)Zhang, Dorner, and Hardt}]{zhang2025benchmark}
Guanhua Zhang, Florian~E. Dorner, and Moritz Hardt. 2025.
\newblock \href {https://doi.org/10.52202/085713-4528} {How benchmark
  prediction from fewer data misses the mark}.
\newblock In \emph{Advances in Neural Information Processing Systems}.

\bibitem[{Zhang et~al.(2026)Zhang, Guo, Lu, Dai, Xia, and
  Wang}]{zhang2026sparseeval}
Taolin Zhang, Hang Guo, Wang Lu, Tao Dai, Shu-Tao Xia, and Jindong Wang. 2026.
\newblock \href
  {https://proceedings.iclr.cc/paper_files/paper/2026/hash/b059de4629cccb411ef92d0d4157e87e-Abstract-Conference.html}
  {{SparseEval}: Efficient evaluation of large language models by sparse
  optimization}.
\newblock In \emph{International Conference on Learning Representations}.

\end{thebibliography}

\appendix
\setcounter{dbltopnumber}{3}
\renewcommand{\dbltopfraction}{0.9}
\renewcommand{\textfraction}{0.1}
\section{Implementation and Source-Only Configuration}
\label{sec:implementation}
\begin{table}[H]\centering\scriptsize
\begin{tabular}{llrrrrr}\toprule
Bank & Fold & $r$ & $\gamma$ & $\kappa$ & Policy & $\lambda$ \\ \midrule
Lite & 0 & 1 & 4 & 1 & Score & 0.5 \\
Lite & 1 & 4 & 4 & 4 & Score & 0.5 \\
Lite & 2 & 4 & 8 & 0.25 & Score & 0.5 \\
Lite & 3 & 1 & 8 & 4 & Latent & 0 \\
Verified & 0 & 4 & 1 & 1 & Score & 0 \\
Verified & 1 & 1 & 8 & 1 & Score & 0.5 \\
Verified & 2 & 1 & 8 & 4 & Latent & 0.5 \\
Verified & 3 & 2 & 4 & 1 & Score & 0.5 \\
Full & 0 & 0 & 0.25 & 1 & Score & 0 \\
Full & 1 & 2 & 4 & 1 & Score & 0 \\
Full & 2 & 4 & 4 & 1 & Score & 0.5 \\
Full & 3 & 4 & 8 & 1 & Score & 0 \\
Pro & 0 & 4 & 1 & 4 & Score & 0 \\
Pro & 1 & 2 & 8 & 16 & Latent & 0 \\
Pro & 2 & 2 & 4 & 4 & Score & 0 \\
Pro & 3 & 4 & 4 & 16 & Score & 0.5 \\
\bottomrule\end{tabular}
\caption{Frozen source-selected settings. For $r>0$, the scalar ability precision is .25 and residual precision is $\gamma$; for $r=0$, ability precision is $\gamma$. Score and Latent denote the acquisition rules in Section~\ref{sec:acquisition}.}\label{tab:hspr-settings}
\end{table}

\paragraph{Item initialization.}
For task $i$ let $h_i=\sum_s M_{si}$. Source-only marginal likelihood fits
a mixture $\pi\delta_0+(1-\pi)\operatorname{Beta}(\alpha,\beta)$ to historical
task rates. For $h_i>0$, the initial rate is
$(h_i+\alpha)/(m+\alpha+\beta)$. For a zero column it is multiplied by
the posterior probability of the continuous component. Rates are clipped
to $[10^{-4},1-10^{-4}]$ before taking their logits.
The zero mass is estimated continuously from source responses.

A source ability is the logit of its Jeffreys-smoothed full score,
standardized using the source mean and a standard deviation floored at .25.
Sixteen regularized Newton updates fit task intercepts and positive
slopes. Intercepts have unit quadratic regularization around their initial
logits; slopes have precision two around one. Updates are clipped to .5
in magnitude, intercepts to $[-10,10]$, and slopes to $[.05,5]$.
These numerical settings are fixed for source fitting.

\paragraph{Residual factors.}
Let $\hat p_{si}=\sigma(b_i+a_i\theta_s)$ be the fitted source probability,
and let $E$ denote $M-\hat p$ centered across sources. Define the mean
fitted Bernoulli variance as
$\bar d_i=m^{-1}\sum_{s=1}^m\hat p_{si}(1-\hat p_{si})$.
For $E=U\Sigma V^\top$, residual loading column $k$ is
\begin{equation}
 v_{ik}=\frac{V_{ik}\Sigma_{kk}}
 {\sqrt{\max(1,m-1)}\max(\bar d_i,.08)}.
\end{equation}
The item model and residual factors are refitted whenever a validation row is removed.
At most eight directions are computed; the residual-rank grid is
$r\in\{1,2,4\}$ with precision $\gamma\in\{1,4,8\}$, plus scalar
$r=0$ with precision $\gamma\in\{.25,1\}$.
For $r>0$, the ability precision is .25 and residual coordinates have
precision $\gamma$. For $r=0$, ability precision is $\gamma$.
Source-signature groups average the intercepts, slopes, and residual
directions, avoiding numerical differences among source-identical items.

\paragraph{Source-only configuration grid.}
HSPR uses staged leave-one-source-agent-out validation over finite
configuration grids. The global grid combines the eleven rank/precision
pairs above with logistic score-variance or latent-state acquisition
and calibration $\lambda\in\{0,.5\}$ (44 settings). This stage retains
the selected rank and precision. With these held fixed, the hierarchical grid uses
$\kappa\in\{.25,1,4,16\}$, Score or Latent acquisition,
and posterior-mean calibration $\lambda\in\{0,.5\}$. The selected
group uncertainty strength and acquisition rule are then held fixed while
median calibration is chosen on the resulting query paths.

The global stage uses deterministic query paths and averages errors over
held-out source agents. The hierarchical and median stages also average
over query seeds 0, 1, and 2. Ties in source MAE use the largest per-agent
validation error, averaged across seeds when applicable, then fixed grid order.
Each held source is excluded before refitting
the response model and task groups. We refit the model on all source
agents using the selected configuration and freeze it before target replay.
Table~\ref{tab:hspr-settings} lists the deployed settings for every fold.

\paragraph{Posterior proposal and updates.}
An auxiliary logistic-normal group MAP supplies a Gaussian proposal for
$z$. Its local offsets have quadratic precision $\kappa$ and are solved
alongside the global coordinates using a Schur complement. The proposal
is refreshed after each eight observed outcomes. Between refreshes, conjugate
updates reweight the existing nodes and update observed counts exactly
within the weighted-node model.

If node $z_j$ is drawn from proposal density $q$, its normalized weight
satisfies
\begin{equation}
 w_j\propto
 \frac{p(z_j)}{q(z_j)}\prod_g
 \frac{\mathrm B(A_g(z_j),F_g(z_j))}
 {\mathrm B(c_gp_g,c_g(1-p_g))}.
\end{equation}
Gaussian normalization constants common to the nodes cancel.
The implementation computes Beta-integrated ordered likelihoods with
rising-factorial \texttt{log1p} terms to avoid cancellation when
concentrations are large. Probabilities used by that likelihood are
clipped to $[10^{-12},1-10^{-12}]$.

The 8,192 nodes use a scrambled Sobol sequence mapped through a standard
normal inverse CDF, with simulation seed 20261005. The predictive median
uses 131,072 complete finite-score draws. Predictive decision
sampling uses the same fixed numerical seed but a separate generator
from the query RNG. Separate generators keep predictive sampling
independent of within-group query randomization. Query seeds 0--4
supply the five target paths.

\section{Derivation of the Acquisition Criterion}
\label{sec:gain-proof}
Let $m_g(z)=A_g/(A_g+F_g)$ and
$\tau_g^2(z)=m_g(z)(1-m_g(z))/(A_g+F_g+1)$.
Conditional on $z,D_t$,
\begin{align}
 \E[K_g\mid z,D_t]&=U_gm_g(z),\\
 \operatorname{Var}(K_g\mid z,D_t)
 &=U_gm_g(z)(1-m_g(z))\notag\\
 &\quad+U_g(U_g-1)\tau_g^2(z).
\end{align}
Integrating the conditional variance and adding the variance of the
conditional score mean gives the predictive finite-score variance.
Both the Bernoulli outcome variance and the shared rate uncertainty
remain in this calculation.

For a particular unqueried item in group $g$, the conditional covariance
with $S$ is
\begin{align}
 \operatorname{Cov}(S,Y_g\mid z,D_t)
 &=\frac{m_g(z)(1-m_g(z))}{N}\notag\\
 &\quad+\frac{(U_g-1)\tau_g^2(z)}{N}.
\end{align}
Marginalizing over $z$ takes the posterior expectation of this conditional
covariance and adds the covariance between the conditional score mean
and $m_g(z)$, with $D_t$ held fixed. This additional term captures global
transfer between groups. Omitting the direct item term would instead
optimize information about other unqueried tasks.

For any random variable $T$ and binary query outcome $Y$ with
$p=\Pr(Y=1\mid D)\in(0,1)$, let
$\mu_y=\E[T\mid D,Y=y]$. Then
\begin{align}
 \operatorname{Cov}(T,Y\mid D)&=p(1-p)(\mu_1-\mu_0),\\
 \operatorname{Var}(\E[T\mid D,Y]\mid D)
 &=p(1-p)(\mu_1-\mu_0)^2.
\end{align}
The law of total variance therefore yields
\begin{align}
 &\operatorname{Var}(T\mid D)
 -\E[\operatorname{Var}(T\mid D,Y)\mid D]\notag\\
 &\qquad=\frac{\operatorname{Cov}(T,Y\mid D)^2}{p(1-p)}.
\end{align}
Taking $T=S$ proves Proposition~\ref{prop:score-gain}; summing coordinatewise
with $T=z_k$ gives Equation~\ref{eq:latentgain}. This identity is exact
for the assumed distribution, including a finite weighted-node
approximation.

\section{Evaluation Protocol and Reproducibility}
\label{sec:accounting}
Algorithm~\ref{alg:hspr} operates on a fixed task bank and reveals target
outcomes through its execution oracle. During source validation, the
entire held-out row is removed before fitting initialization rates,
item parameters, residual factors, and source signatures. The model is
then refitted on all source rows using the selected configuration.
All model, query-policy, and calibration choices are frozen before
target replay.

At each step, acquisition uses sources and observed history only. Exhausted
groups are excluded, and one remaining task is sampled uniformly within
the selected group before revealing its binary outcome. Every path
contains $B$ distinct executions. Each method and query path has its own
$B$-execution budget. Predictive draws use a numerical generator separate
from query selection.

The score estimate and queried task identities are recorded before full
target outcomes are accessed. Full truth then supplies MAE, ranking,
and maximum error. The complete target folds are pooled within each
query seed before metrics are averaged, as defined in
Section~\ref{sec:protocol}. This ordering separates inference information
from scoring information. Historical bank collection and source fitting
are additional costs outside $B$.

\paragraph{Adapted baseline protocols.}
FAQ-port uses the without-replacement variant and caps its logistic factor
rank at the source cohort size minus one. It retains posterior factor
updates and the PAI estimator; this variant does not inherit the
with-replacement coverage theorem. ATLAS-port retains sequential
expected-a-posteriori ability estimation, maximum-Fisher-information
selection with top-five randomization, and p-IRT score estimation.
It uses the source-fold 4PL bank in place of the original R/mirt 3PL
fitting and applies fixed distinct-task budgets. Thus these rows report
adapted procedures on the available banks.
RSL uses the Uniform query set; Mid-Range raw and Mid-Range with Ridge
share their source-selected task set. Neither shares HSPR's adaptive
query paths. PTA-IRT retains its released frozen predictions, including
its use of historical trajectory information. The matched target-task
budget does not equalize historical information or source-fitting costs.

\section{Complete Paired Statistical Comparisons}
\label{sec:paired-comparisons}
\begin{table*}[!tbp]\centering\small
\begin{tabular}{llrrr}\toprule
Benchmark & Comparator & $\Delta$ MAE (pp) & Paired 95\% CI (pp) & Holm $p$ \\ \midrule
Lite & ATLAS-port & -0.762 & $[-1.370, -0.162]$ & 0.081 \\
Lite & PTA-IRT & -0.797 & $[-1.724, +0.006]$ & 1.000 \\
Lite & Stratified & -2.541 & $[-3.205, -1.886]$ & $<.001$ \\
Lite & RSL & -2.984 & $[-3.812, -2.258]$ & $<.001$ \\
Verified & ATLAS-port & +0.055 & $[-0.355, +0.448]$ & 1.000 \\
Verified & PTA-IRT & -0.262 & $[-0.825, +0.283]$ & 1.000 \\
Verified & Stratified & -1.387 & $[-1.803, -0.955]$ & $<.001$ \\
Verified & RSL & -1.738 & $[-2.240, -1.257]$ & $<.001$ \\
Full & ATLAS-port & -0.240 & $[-0.645, +0.132]$ & 1.000 \\
Full & PTA-IRT & -1.562 & $[-2.631, -0.588]$ & 0.116 \\
Full & Stratified & -0.919 & $[-1.132, -0.696]$ & 0.001 \\
Full & RSL & -1.559 & $[-2.714, -0.681]$ & 0.017 \\
Pro & ATLAS-port & -5.638 & $[-8.068, -3.313]$ & 0.003 \\
Pro & PTA-IRT & -0.633 & $[-2.472, +0.721]$ & 1.000 \\
Pro & Stratified & -1.012 & $[-1.676, -0.315]$ & 0.122 \\
Pro & RSL & -2.369 & $[-4.105, -1.033]$ & 0.036 \\
\bottomrule\end{tabular}
\caption{Paired MAE differences: HSPR minus comparator, so negative favors HSPR. Seed-averaged agent errors are paired. The 95\% percentile bootstrap intervals use 10,000 configuration resamples, conditional on the archived task bank, and are unadjusted. Wilcoxon $p$-values are Holm-adjusted across all 16 comparisons.}\label{tab:hspr-paired}
\end{table*}

The contrasts in Table~\ref{tab:hspr-paired} all use the same held-out
configuration keys within a benchmark. Baseline seed counts differ,
so seed-averaged errors supply one paired observation per configuration.
The bootstrap measures variability over the archived configuration sample,
conditional on the observed task bank. Wilcoxon tests use the same global
16-contrast Holm correction as Section~\ref{sec:results}.

\section{Additional Controlled Analyses and Numerical Validation}
\label{sec:numerical-checks}
\paragraph{End-to-end hierarchy controls.}
Table~\ref{tab:hspr-controls}A retains each fold's source-selected acquisition
objective (13 Score and three Latent folds), global rank and precision,
and adopted median calibration. Removing hierarchy sets
the group rate to its global response probability, eliminating the
conditional Beta variability. Each control updates its own posterior
and selects its own distinct-task path. Both retain all 133 target
configurations and five query seeds. The equal-bank macro difference
uses 10,000 paired configuration bootstrap samples within each bank,
after averaging seeds. Removing hierarchy increases macro MAE from
1.664 to 1.808 pp; the 0.143 pp difference has a conditional 95\%
interval of $[-0.012,0.302]$ pp, which includes zero.
Individual-bank Wilcoxon tests do not remain significant after correcting
the four hierarchy comparisons with Holm's method. The macro interval summarizes aggregate
variation across the archived configuration cohorts.

\paragraph{Predictive precision.}
Source validation checks that predictive sample means
are within six Monte Carlo standard errors of analytic posterior means.
For the median's numerical interval, binomial order-statistic quantiles
identify the central 95\% indices among 131,072 sorted score draws.
Its width is checked against one finite task increment $1/N$.
This interval measures Monte Carlo precision of the posterior median.

\paragraph{Integration resolution.}
Source-only checks compare 8,192 and 32,768 nodes on common query
histories, with a score tolerance of .0005 on the 0--1 scale (.05 pp).
The Verified draw-resolution check compares 65,536 and 131,072 draws
using the same frozen query policy and factor convention.
These comparisons assess numerical accuracy at the reported operating
settings.

\paragraph{Matched-path reconstruction.}
\begin{table}[H]\centering\scriptsize
\setlength{\tabcolsep}{2.5pt}
\begin{tabular}{lrrrrrr}\toprule
Bank & $N$ & $\bar G$ & $N/\bar G$ & \multicolumn{2}{c}{MAE $\downarrow$} & \shortstack{Group\\gain $\uparrow$} \\
\cmidrule(lr){5-6}
 & & & & \shortstack{w/o\\groups} & \shortstack{with\\groups} & \\ \midrule
Lite & 300 & 201.5 & 1.49 & 1.949 & 1.941 & +0.009 \\
Verified & 500 & 426.0 & 1.17 & 1.928 & 1.927 & +0.001 \\
Full & 2294 & 275.9 & \textbf{8.32} & 0.902 & 0.722 & \textbf{+0.180} \\
Pro & 730 & 222.0 & 3.29 & 2.101 & 2.108 & -0.006 \\
\bottomrule\end{tabular}
\caption{Historical group structure and matched-path reconstruction at 10\% budgets. Both MAE columns use posterior means with identical query paths and calibration. $\bar G$ is the target-averaged source group count. Group gain is MAE without minus with group uncertainty, computed before rounding (pp). Bold marks the largest $N/\bar G$ and gain.}\label{tab:hspr-matched-path}
\end{table}

The reported same-path no-group control retains the same global
configuration and selected tasks while integrating the global posterior
without a Beta rate residual. It is evaluated as a posterior mean.
Comparing it with the hierarchical mean separates the effect of
conditional group uncertainty on these paths. Table~\ref{tab:hspr-controls}C
compares the posterior mean with the median on identical query paths,
with calibration selected separately for each decision family.
Table~\ref{tab:hspr-matched-path} reports the complete matched-path
comparison. Its effect varies across banks: the largest improvement is
on Full, while Pro slightly favors the no-group mean.

\paragraph{Acquisition controls.}
Table~\ref{tab:hspr-controls}B retains all HSPR model settings and median
calibration. Uniform reuses saved baseline permutations for seeds
11, 23, 37, 53, and 71; Mid-Range sorts source rates by distance to .5
with stable task-order ties. All 133 targets yield 665 Uniform and
133 deterministic Mid-Range predictions.

Thirty-two source-only checks compare integration resolutions on
identical paths, excluding held sources before fitting and selection.
Predictive sampling for the static acquisition controls uses a
predefined numerical-precision rule: start at 131,072 draws and double,
up to 2,097,152, until sample means agree within six Monte Carlo standard
errors and median numerical intervals span at most $1/N$.
This rule uses numerical error estimates before full target scoring.
All adaptive references meet the criteria; 797 static predictions use
131,072 draws and one uses 524,288.
Recorded static paths reproduce their baseline selectors.

The budget sweep preserves the primary 10\% predictions and checks
all source checkpoints at the same numerical tolerances.

\section{Historical Response Structure and Transfer Diagnostics}\label{sec:support-diagnostic}
\begin{table}[H]\centering\scriptsize
\setlength{\tabcolsep}{2.5pt}
\begin{tabular}{lrrrr}\toprule
Bank & \shortstack{Source\\groups} & \shortstack{Source-zero\\tasks (\%)} & \shortstack{Target success\\on zeros (\%)} & \shortstack{Queries to\\zeros (\%)} \\ \midrule
Lite & 201.5 & 24.6 & 0.87 & 0.248 \\
Verified & 426.0 & 8.3 & 0.74 & 0.000 \\
Full & 275.9 & 41.3 & 1.84 & 13.225 \\
Pro & 222.0 & 36.7 & 2.07 & 0.020 \\
\bottomrule\end{tabular}
\caption{Post-hoc support diagnostic at 10\% budgets. Each target uses its frozen source fold; entries average targets equally, with queries averaged over five seeds. Source-zero tasks have zero successes among all source agents. Target success rates are computed after prediction for this diagnostic.}\label{tab:hspr-support}
\end{table}

Table~\ref{tab:hspr-support} gives the complete per-bank support
statistics underlying Section~\ref{sec:hspr-mechanism}. Entries average
configurations equally, using each target's frozen source fold.
Target success is measured within that target's source-zero tasks;
query coverage is the fraction of its queried tasks drawn from this set,
averaged over five seeds. Query coverage reflects the deployed
source-selected acquisition policy.

\section{Computational Overhead}
\label{sec:compute-overhead}
We profiled HSPR at the 10\% task budget over all 133 target configurations
and five query seeds (665 runs). Experiments used an AMD EPYC 7352 server
with 96 logical CPUs and approximately 566 GiB of system memory, running
Python 3.11.3, NumPy 1.26.4, and SciPy 1.17.1. We ran 24 evaluations
concurrently, each in a separate process with one CPU thread, without a GPU.
Timing includes posterior initialization, sequential querying and updates,
and the final predictive decision and calibration. It excludes data loading,
source fitting and configuration selection, and software-agent task execution.

\begin{table}[t]
\centering
\small
\begin{tabular}{lrrr}
\toprule
Task bank & Runs & Median (s) & P90 (s) \\
\midrule
Lite & 175 & 6.19 & 8.57 \\
Verified & 350 & 20.31 & 32.35 \\
Full & 70 & 30.96 & 61.67 \\
Pro & 70 & 7.79 & 14.12 \\
\bottomrule
\end{tabular}
\caption{HSPR evaluator wall time per target and query seed at the 10\% task budget. P90 is the 90th percentile across runs within each bank.}
\label{tab:compute-overhead}
\end{table}

Table~\ref{tab:compute-overhead} reports per-run wall times on a shared
server. The complete batch took 510.1 seconds with concurrent evaluation;
the timed inference intervals consumed 3.31 CPU hours in total.
All replayed query paths matched the archived primary results, and the
largest absolute prediction difference was $2.22\times10^{-16}$ on the
unit score scale. These measurements quantify deployment inference
overhead, not the total computational cost of source-only configuration
selection or end-to-end agent evaluation savings.

\section{Score Recovery Across Model Families}
\label{sec:configuration-estimates}
\begin{figure*}[t]
\centering
\includegraphics[width=\textwidth]{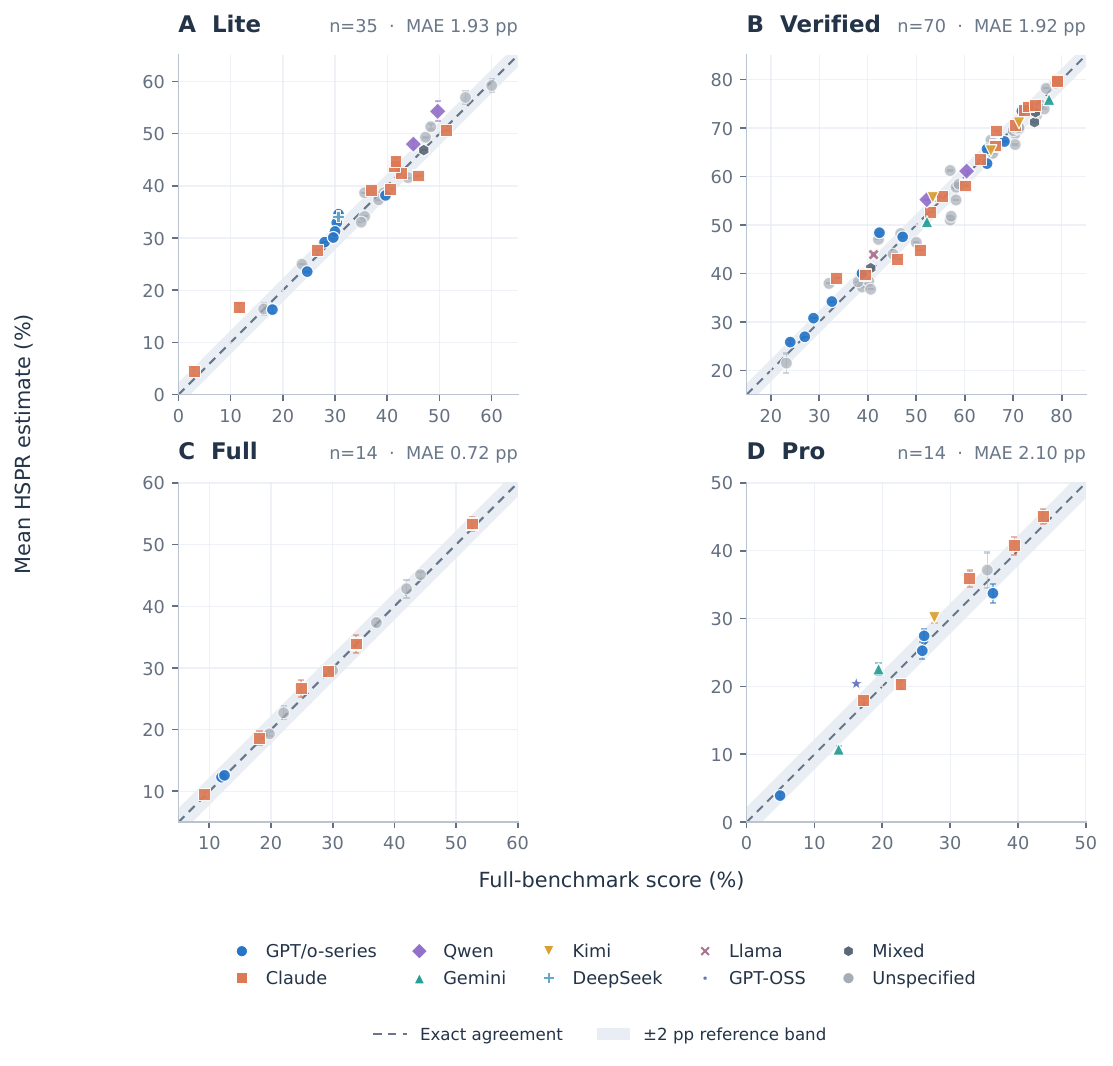}
\caption{Full-benchmark scores and HSPR estimates by model family at the 10\% task budget. Points show mean predictions over five query seeds; bars show one sample standard deviation across those seeds. The dashed line marks exact agreement, and the shaded reference band marks $\pm2$ pp. Panel MAE averages absolute errors over configurations and seeds. Family labels follow archived configuration names; Mixed denotes multiple named families and Unspecified denotes no identified family.}
\label{fig:configuration-estimates}
\end{figure*}

Figure~\ref{fig:configuration-estimates} compares full-benchmark scores with
HSPR estimates at the 10\% task budget. Each point represents one target
configuration, with model families distinguished by color and marker.
The diagonal shows exact agreement, making overestimation and
underestimation visible across the score range.

The relative errors of the represented families vary across task banks.
On Lite, GPT/o-series configurations have lower mean error than Claude
configurations (1.71 versus 2.15 pp), while the ordering reverses on
Verified (1.70 versus 1.52 pp). Full shows consistently small errors
across its configurations: the GPT/o-series, Claude, and unspecified
groups have mean errors of 0.39, 0.80, and 0.74 pp, respectively, and
every configuration's mean absolute error is below 1.71 pp. The individual
points also reveal variation among configurations within a family that
is hidden by family averages.

\end{document}